\documentclass[runningheads]{llncs}

\usepackage{eccv}

\usepackage{eccvabbrv}

\usepackage{graphicx}
\usepackage{booktabs}

\usepackage[accsupp]{axessibility}  

\usepackage{amsmath}
\DeclareMathOperator*{\argmin}{arg\,min}
\usepackage[ruled,linesnumbered]{algorithm2e}
\usepackage{ulem}  

\usepackage{multirow}
\usepackage{xcolor}
\usepackage{colortbl}
\usepackage{array}

\definecolor{cvprblue}{HTML}{2563EB}
\definecolor{secdark}{HTML}{222222}
\definecolor{green_spd}{HTML}{16A34A}
\definecolor{gray_base}{HTML}{999999}
\definecolor{gray_arrow}{HTML}{666666}
\usepackage[table]{xcolor}
\usepackage{colortbl}
\definecolor{headerbg}{HTML}{E9EDF5}
\definecolor{rowbg}{HTML}{F7F8FB}
\definecolor{anno}{HTML}{999999}

\usepackage{enumitem}

\newcommand{\tablestyle}[2]{\setlength{\tabcolsep}{#1}\renewcommand{\arraystretch}{#2}}
\newcommand{\shline}{\noalign{\global\arrayrulewidth=0.8pt}\hline\noalign{\global\arrayrulewidth=0.4pt}}
\newcommand{\fps}[3]{%
  {\color{gray_base}#1}{\color{gray_arrow}$\rightarrow$}{\color{cvprblue}\textbf{#2}} {\color{gray_arrow}(}{\color{green_spd}\textbf{#3$\times$}}{\color{gray_arrow})}%
}
\newcommand{\dash}{{\color{gray_base}---}}

\usepackage{hyperref}

\usepackage{orcidlink}

\begin{document}

\title{Fast SAM 3D Body: Accelerating SAM 3D Body for Real-Time Full-Body Human Mesh Recovery} 

\titlerunning{Fast SAM 3D Body}

\author{Timing Yang\inst{1} \and
Sicheng He\inst{1} \and
Hongyi Jing\inst{1} \and
Jiawei Yang\inst{1} \and
Zhijian Liu\inst{2,3} \and
Chuhang Zou\inst{4}\textsuperscript{\normalsize$\dagger$} \and
Yue Wang\inst{1,3}\textsuperscript{\normalsize$\dagger$}}

\authorrunning{T.~Yang et al.}

\institute{
USC Physical Superintelligence (PSI) Lab \and
University of California, San Diego \and
NVIDIA \and
Meta Reality Labs
}
\maketitle
\renewcommand{\thefootnote}{}
\footnotetext{\textsuperscript{\normalsize$\dagger$} Joint corresponding authors.}
\footnotetext{Code is available at: \url{https://github.com/yangtiming/Fast-SAM-3D-Body}}
\renewcommand{\thefootnote}{\arabic{footnote}}

\begin{center}
    \includegraphics[width=0.9\textwidth]{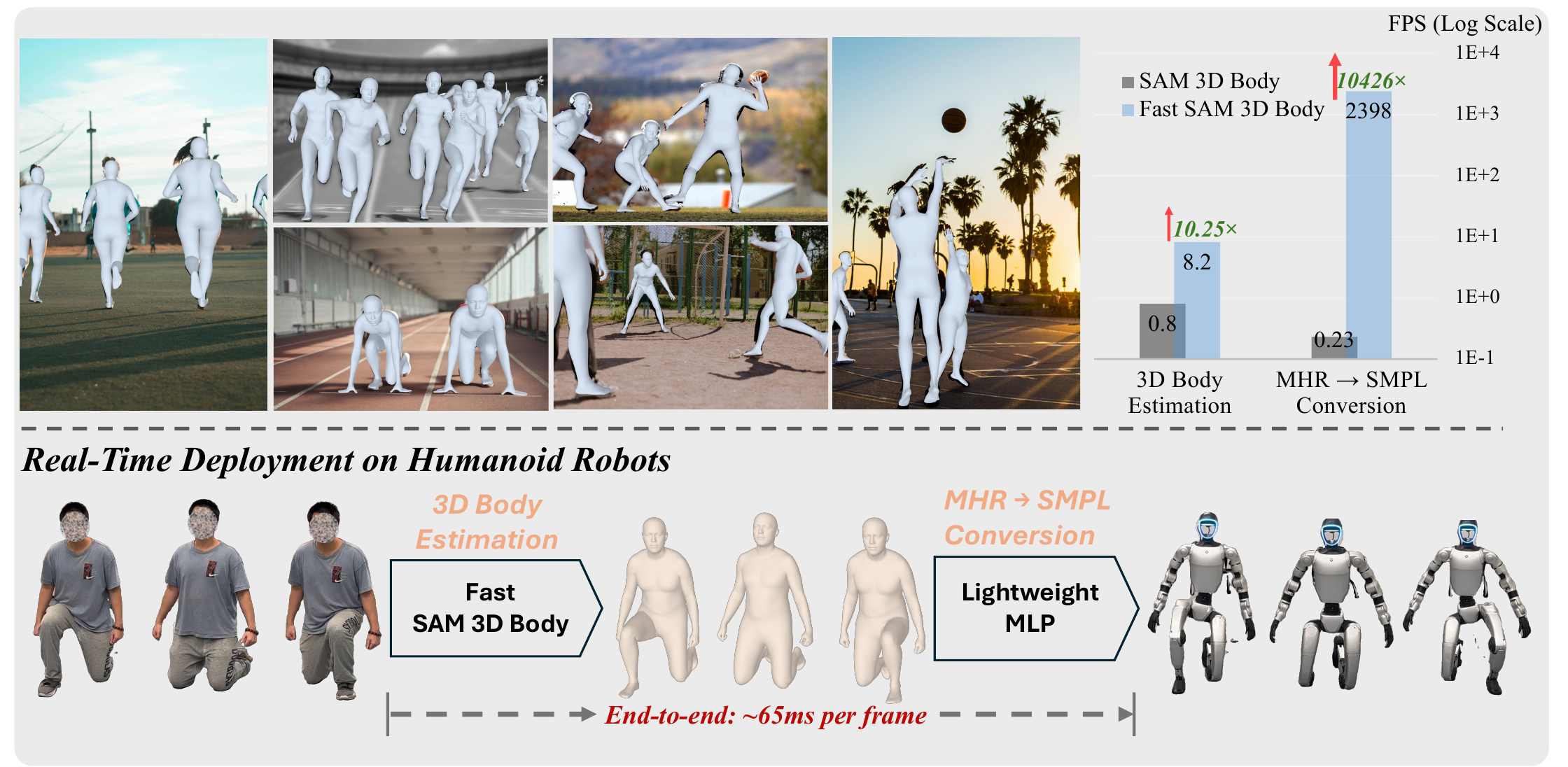}
    \captionof{figure}{
    \textbf{Speed--accuracy overview of Fast SAM 3D Body}. Top left: Qualitative results on in-the-wild images show our framework preserves high-fidelity reconstruction. Top right: Our method achieves up to a $10.25\times$ end-to-end speedup over SAM 3D Body~\cite{sam3db} and replaces the iterative MHR-to-SMPL bottleneck~\cite{MHR:2025} with a $10{,}000\times$ faster neural mapping. Bottom: Our system enables real-time humanoid robot control from a single RGB stream at ${\sim}65$\,ms per frame on an NVIDIA RTX 5090.
    }
    \label{fig:teaser}
\end{center}

\begin{abstract}

SAM 3D Body (3DB) achieves state-of-the-art accuracy in monocular 3D human mesh recovery, yet its inference latency of several seconds per image precludes real-time application. We present \textbf{Fast SAM 3D Body}, a training-free acceleration framework that reformulates the 3DB inference pathway to achieve interactive rates. By decoupling serial spatial dependencies and applying architecture-aware pruning, we enable parallelized multi-crop feature extraction and streamlined transformer decoding. Moreover, to extract the joint-level kinematics (SMPL) compatible with existing humanoid control and policy learning frameworks, we replace the iterative mesh fitting with a direct feedforward mapping, accelerating this specific conversion by over 10,000×. Overall, our framework delivers up to a 10.9× end-to-end speedup while maintaining on-par reconstruction fidelity, even surpassing 3DB on benchmarks such as LSPET. We demonstrate its utility by deploying Fast SAM 3D Body in a vision-only teleoperation system that—unlike methods reliant on wearable IMUs—enables \textbf{real-time humanoid control} and the direct collection of manipulation policies from a single RGB stream.
\end{abstract}

\section{Introduction}
\label{sec:intro}

Monocular 3D human mesh recovery (HMR)—estimating the 3D pose and shape of a person from a single RGB image—is a foundational capability for augmented reality, biomechanics, and human-robot interaction. For these applications, real-time responsiveness is as critical as accuracy; high-fidelity reconstructions are of limited utility if they cannot keep pace with human motion or provide immediate feedback for control loops. While the recent adoption of large vision backbones and expressive body models~\cite{expose2020,smplerx,multihmr2024,pve} has significantly advanced reconstruction accuracy, it has done so at the cost of immense computational overhead.

A prime exemplar of this trade-off is SAM 3D Body (3DB)~\cite{sam3db}. While 3DB achieves state-of-the-art performance, the computational intensity of its multi-stage pipeline—robust detection~\cite{vitdet}, sequential encoding~\cite{dinov3}, and iterative MHR-to-SMPL conversion~\cite{MHR:2025}—bounds inference to sub-FPS speeds. Prior acceleration efforts~\cite{fastmetro2022,potter2023,feater2023,deforhmr2025,tore2023,fasthmr2025,vita2024} primarily redesign ViT architectures. Because these methods optimize only isolated neural components, they are not designed to resolve the cross-stage dependencies and intensive post-processing required by comprehensive pipelines like 3DB, thus falling short of real-time end-to-end speeds. Furthermore, applying such architectural modifications requires extensive retraining, which risks degrading the exceptional generalization ability provided by 3DB's pre-trained backbones. Bridging this gap demands a holistic reformulation of the inference pathway—one that streamlines these compound dependencies to unlock real-time performance through a training-free approach.

To this end, we present \textbf{Fast SAM 3D Body}, a training-free acceleration framework that holistically reformulates the 3DB pipeline. As shown in Fig.~\ref{fig:teaser}, rather than redesigning the model architecture, we preserve 3DB's robust generalization ability while unlocking real-time performance through three methodological shifts:
(i) \textbf{Spatial Dependency Decoupling:} We circumvent the baseline's serial body-to-hand decoding bottleneck by introducing a lightweight 2D pose prior~\cite{yolo11}. By analytically deriving extremity bounding boxes from coarse keypoints, we decouple spatial localization from the main decoder, directly unlocking parallelized, multi-crop feature extraction in a single batched forward pass.
(ii) \textbf{Compute-Aware Decoding:} We prune redundant prompt queries and bypass iterative self-refinement. This yields a deterministic execution graph, unlocking low-level hardware compilation (via torch.compile and TensorRT) without sacrificing dynamic expressivity.
(iii) \textbf{Neural Kinematic Projection:} Because standard humanoid control, teleoperation pipelines, and policy learning frameworks widely adopt the SMPL representation, projecting expressive MHR surfaces into this kinematic manifold is a prerequisite for downstream actuation. To bridge this topological gap, we replace the iterative MHR-to-SMPL conversion with a lightweight feedforward mapping, directly projecting MHR features into the actuatable joint space.

In summary, our key contributions are as follows:
\begin{itemize}[leftmargin=*,itemsep=2pt,topsep=4pt]
\item 
A training-free acceleration framework that holistically reformulates the multi-stage SAM 3D Body pipeline, achieving up to a 10.9× end-to-end speedup while maintaining on-par reconstruction fidelity, even surpassing 3DB on benchmarks such as LSPET.
\item 
A learned feedforward MHR-to-SMPL projection module that replaces hundreds of iterative optimization steps, accelerating cross-topology mesh conversion by over 10,000× without compromising millimeter-level precision.
\item 
The successful deployment of this framework in a vision-only, single-RGB teleoperation system that—unlike methods reliant on cumbersome wearable IMUs—enables real-time humanoid robot control and the direct collection of deployable whole-body manipulation policies.
\end{itemize}

\section{Related Work}
\label{sec:Related_Work}

\paragraph{Human Mesh Recovery.}

Monocular HMR has transitioned from body-only regression~\cite{kanazawa2018hmr,hmr2b,cliff2022,tokenhmr2024,camerahmr,prompthmr} to high-fidelity full-body estimation including hands and facial expression~\cite{expose2020,smplerx,multihmr2024,nlf}. While part-specific methods~\cite{pavlakos2024reconstructing,potamias2025wilor} offer specialized accuracy for the extremities, they often lack the holistic context required for global trajectory estimation and temporal coherence~\cite{wham,tram,genmo}. This evolution toward expressive, unified models has necessitated large vision backbones, which—while significantly improving reconstruction quality—has made real-time deployment computationally expensive. 3DB~\cite{sam3db} represents the current state-of-the-art in this trajectory. By leveraging a promptable encoder–decoder architecture and the expressive MHR~\cite{MHR:2025} representation, it achieves unprecedented recovery. However, the rigor of its multi-stage design—incorporating dense detection, sequential multi-crop feature extraction, and iterative kinematic fitting—accumulates latency. Our work resolves these bottlenecks through holistic reformulations that enable real-time execution while retaining 3DB’s robust generalization in a training-free manner.

\paragraph{Parametric Human Body Models.}

SMPL~\cite{smpl} remains a widely adopted representation for HMR, parameterizing pose and shape via learned blend shapes and linear blend skinning. Recently, MHR~\cite{MHR:2025} introduced a decoupled parameterization that separates skeletal structure from soft-tissue deformation. However, because standard evaluation benchmarks and existing robotic control frameworks (e.g., humanoid teleoperation pipelines) are built specifically around the SMPL skeletal topology, models predicting MHR must topologically translate their outputs for practical use. In 3DB, this cross-topology translation is performed via a computationally intensive hierarchical iterative optimization (Eq.~\eqref{eq:mhr2smpl}) that constitutes the primary pipeline bottleneck. We obviate this costly optimization by introducing a lightweight feedforward mapping (Eq.~\eqref{eq:mlp_convert}) that achieves real-time inter-model projection with accuracy on par with iterative fitting.

\paragraph{Efficient Human Mesh Recovery.}
Efforts to reduce HMR inference costs have largely focused on two independent directions. The first is architectural redesign, including lighter encoder–decoders~\cite{fastmetro2022}, efficient attention~\cite{potter2023,feater2023,deforhmr2025}, and token/layer pruning~\cite{tore2023,fasthmr2025,vita2024,tome2023}. While effective, these methods typically address only the vision backbone and require extensive retraining. The second direction is system-level optimization—such as graph compilation and pipeline parallelism~\cite{rtmpose2023,hyperpose2021,expedit2022}—which has proven successful in 2D pose estimation but remains under-explored for multi-stage 3D pipelines. High-fidelity mesh recovery presents a unique "compound latency" profile, where detection, multi-crop encoding, and topology conversion all contribute to sub-real-time rates. Our work unifies these directions, applying holistic optimizations across all inference stages without retraining, thus preserving the "in-the-wild" robustness of the underlying state-of-the-art model.

\section{Method}
\label{sec:Method}

\subsection{Preliminaries: SAM 3D Body (3DB) Pipeline}
\label{sec:prelim}

3DB~\cite{sam3db} recovers full-body 3D human meshes from a single RGB image using a promptable encoder--decoder built on the expressive MHR~\cite{MHR:2025} representation. While highly accurate, 3DB's inference pipeline suffers from structural latencies, which we outline below as the primary targets for our acceleration framework.

\paragraph{Detection and Encoding.} Given an image $\mathbf{I} \in \mathbb{R}^{H \times W \times 3}$, a detector extracts body bounding boxes $\{b_i\}_{i=1}^{N}$, while a field-of-view estimator predicts camera intrinsics $\mathbf{K} \in \mathbb{R}^{3 \times 3}$. Each cropped region $\mathbf{I}_{\mathrm{body}} \in \mathbb{R}^{S \times S \times 3}$ passes through a vision backbone to extract dense features:$$\mathbf{F} = \mathrm{Enc}(\mathbf{I}_{\mathrm{body}}) \in \mathbb{R}^{h \times w \times D}$$ where $h{=}w{=}S/p$, $p$ is the patch size, and $D$ is the embedding dimension. These features are optionally fused with Fourier-encoded pixel rays derived from $\mathbf{K}$.

\paragraph{Tokenized Transformer Decoding.}The body decoder processes a concatenated sequence of learnable query tokens representing the initial MHR state, spatial prompts, and keypoints:$$\mathbf{T} = [\, \mathbf{t}_{\mathrm{mhr}},\; \mathbf{T}_{\mathrm{prompt}},\; \mathbf{T}_{\mathrm{kp2d}},\; \mathbf{T}_{\mathrm{kp3d}},\; \mathbf{T}_{\mathrm{hand}}\,] \in \mathbb{R}^{M \times D}$$During the $L$-layer decoding, these tokens cross-attend to the image features $\mathbf{F}$. A defining characteristic of 3DB is its intermediate prediction mechanism: after every layer $\ell$, the output token $\mathbf{t}^{(\ell)}_{\mathrm{mhr}}$ decodes intermediate MHR parameters and camera estimates. Forward kinematics yields 3D joints $\mathbf{J}^{(\ell)}$ and projected 2D keypoints $\hat{\mathbf{J}}_{2\mathrm{d}}^{(\ell)}$, which dynamically update the positional encodings for the subsequent layer:$$\mathbf{P}_{\mathrm{kp2d}}^{(\ell+1)} = \phi_{\mathrm{2d}}\!\bigl(\hat{\mathbf{J}}_{2\mathrm{d}}^{(\ell)}\bigr), \qquad \mathbf{P}_{\mathrm{kp3d}}^{(\ell+1)} = \phi_{\mathrm{3d}}\!\bigl(\mathbf{J}^{(\ell)} - \bar{\mathbf{J}}^{(\ell)}\bigr)$$where $\phi_{\mathrm{2d}}, \phi_{\mathrm{3d}}$ are learned linear projections and $\bar{\mathbf{J}}^{(\ell)}$ is the pelvis location.

\paragraph{Sequential Hand Decoding and Refinement.}
Following body decoding, hand bounding boxes are predicted by a dedicated head. Each hand crop is independently encoded and decoded to yield refined hand MHR parameters, which are merged with the body. To ensure joint alignment, the merged 2D keypoints are fed back as spatial prompts for a complete second forward pass through the body decoder.

\paragraph{Iterative MHR-to-SMPL Conversion.}
Because standard benchmarks evaluate in the SMPL~\cite{smpl} topology, the predicted MHR mesh undergoes a hierarchical iterative optimization~\cite{MHR:2025}:
\begin{equation}
    \hat{\boldsymbol{\Theta}}_{\mathrm{smpl}} = \argmin_{\boldsymbol{\Theta}}
    \bigl\|\mathbf{V}_{\mathrm{mhr}} - \mathbf{V}_{\mathrm{smpl}}(\boldsymbol{\Theta})\bigr\|^2
    + \mathcal{R}(\boldsymbol{\Theta}),
    \label{eq:mhr2smpl}
\end{equation}
where $\mathbf{V}_{\mathrm{mhr}}$ is the MHR mesh predicted by the decoder, $\mathbf{V}_{\mathrm{smpl}}(\boldsymbol{\Theta})$ is the SMPL mesh, and $\mathcal{R}$ comprises anatomical regularizers. This cross-topology fitting requires hundreds of steps per person.

\begin{figure}[!t]
  \centering
  \includegraphics[width=\linewidth]{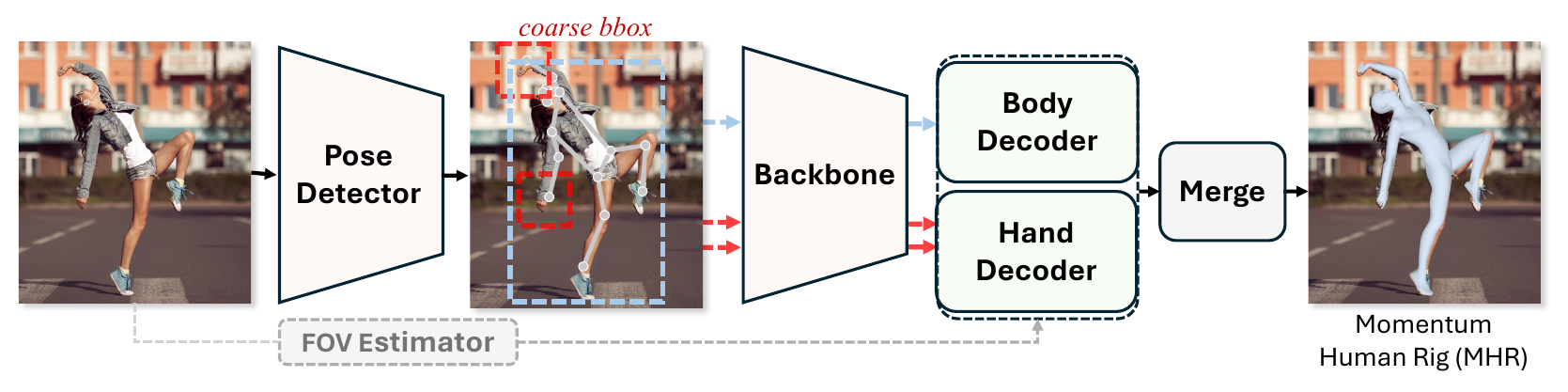}
  \vspace{-7mm}
  \caption{\textbf{\textit{Overview of the Fast SAM 3D Body pipeline.}} 
A pose detector predicts 2D body keypoints from which body and hand 
coarse bboxes are derived simultaneously, enabling all crops to be encoded in a single batched backbone pass. A lightweight FOV estimator predicts camera intrinsics. The body and hand decoders process the resulting features, and their outputs are merged to produce the final 
MHR mesh.}
\vspace{-2mm}
  \label{fig:pipeline}
\end{figure}

\subsection{Acceleration}
\label{sec:accel}

We address the compound latencies of the 3DB pipeline through a series of algorithmic reformulations. Rather than treating pipeline stages in isolation, we holistically resolve cross-stage dependencies, transform dynamic execution into deterministic graphs, and bypass optimization bottlenecks. The reformulations targeting the 3DB inference pathway itself require no weight changes; the MHR-to-SMPL conversion \cite{MHR:2025}---independent of 3DB itself and needed only for downstream kinematic control---is instead replaced with a learned topological projection, leaving all 3DB weights frozen. Fig.~\ref{fig:pipeline} illustrates our streamlined pipeline, and Fig.~\ref{fig:pipeline_comparison} provides a detailed stage-by-stage comparison with the original 3DB.

\subsubsection{Decoupled Spatial Priors}\label{sec:detection}In the original 3DB architecture, hand detection is intrinsically entangled with the body decoder, creating a major serial dependency: hand crops cannot be spatially resolved until the body decoding concludes. We circumvent this bottleneck by introducing a lightweight spatial prior. By predicting coarse 2D body keypoints via a single-stage pose detector in an initial pass, we analytically derive the bounding boxes for extremities prior to any transformer execution. Specifically, the hand bounding box $b_{\mathrm{hand}}$ is deterministically computed as:$$b_{\mathrm{hand}} = \bigl[\,x_w - \tfrac{s}{2},\; y_w - \tfrac{s}{2},\; x_w + \tfrac{s}{2},\; y_w + \tfrac{s}{2}\,\bigr],$$where $(x_w, y_w)$ is the predicted wrist keypoint location and $s = \min(w_{\mathrm{body}}, h_{\mathrm{body}}) / \alpha$ is a scale factor derived from the global body dimensions. Crucially, because the high-capacity downstream decoder is robust to minor spatial shifts, these initial keypoints do not require sub-pixel precision; they merely need to establish a reliable bounding region. This decoupling entirely abstracts the spatial preparation of extremities away from the main decoder, directly enabling batched multi-crop feature extraction while drastically reducing initial detection latency.

\subsubsection{Static Graph Reformulation}
\label{sec:backbone}
A critical bottleneck in expressive HMR pipelines is the dynamic, data-dependent nature of their execution. 3DB's reliance on asynchronous crop generation and dynamic token updating creates an unpredictable computation graph, incurring severe kernel launch and memory allocation overheads. We restructure the inference pathway into a strict, static execution graph. Depending on the target hardware ecosystem, we utilize two parallel compilation strategies: (i)~compiling the backbone into a TensorRT FP16 engine with dynamic batching, which fuses operations and executes asynchronously; or (ii)~leveraging native static compilation to capture the forward pass as a CUDA Graph, allowing for deterministic replay with near-zero launch overhead. We apply an identical compilation strategy to the FOV estimator, purposely selecting its most compact model variant at the lowest resolution, as our empirical ablations indicate the FOV task saturates early at this capacity.

\subsubsection{Compute-Aware Decoder Pruning}
\label{sec:decoder}
The transformer decoder is the most complex component to optimize, involving the interplay of cross-attention layers, intermediate predictions, and feedback loops. We apply several complementary theoretical and structural strategies.

\paragraph{Intermediate prediction pruning.}
The original decoder executes a full intermediate prediction (IntermPred) after every layer $\ell < L$, requiring costly forward kinematics, skinning, and camera projection each time. However, we posit that early transformer layers primarily capture low-level semantic correlations, making full kinematic projection mathematically redundant at these initial stages. To exploit this, we introduce a configurable layer selection set $\mathcal{S} \subset \{0, \dots, L{-}1\}$ that strictly gates this execution. For any layer $\ell \notin \mathcal{S}$, the expensive IntermPred is bypassed entirely; the keypoint token positional encodings $\mathbf{P}_{\mathrm{kp2d}}^{(\ell+1)}$ and $\mathbf{P}_{\mathrm{kp3d}}^{(\ell+1)}$ simply retain their cached values from the most recent update, preserving the feedback signal without redundant recomputation.

\paragraph{Disabling keypoint-prompted refinement.}
The self-prompting refinement pass, in which predicted 2D keypoints are fed back as prompts for a second decoder forward pass, is removed at inference. Because our spatial priors are already accurately decoupled, eliminating this entire additional decoder evaluation results in no measurable degradation in reconstruction accuracy.

\subsubsection{Pipeline Restructuring}
\label{sec:pipeline}
The original pipeline processes body and hand crops sequentially through three independent backbone forward passes, with hand crops prepared on the CPU. We consolidate this into a single batched pass

\paragraph{GPU-native hand crop preprocessing.}
We eliminate the GPU--CPU--GPU round trip per hand by constructing differentiable sampling grids directly from the analytically derived bounding box coordinates. Both crops are extracted in a single bilinear interpolation pass on the GPU, removing data transfer overhead.

\begin{figure}[t]
\centering
\tiny
\setlength{\tabcolsep}{3pt}
\renewcommand{\arraystretch}{1.0}
\begin{tabular}{@{}l@{\;\;}p{0.41\linewidth}|p{0.41\linewidth}@{}}
\toprule
\rowcolor{headerbg}
& \textbf{Original (SAM 3D body)} & \textbf{Ours (Fast SAM 3D body)} \\
\midrule
\rowcolor{headerbg}
\multicolumn{3}{@{}l}{\textbf{Input:} Image $\mathbf{I}$ \quad\textbf{Output:} MHR parameters $\hat{\theta}$\; {\color{anno}(+ SMPL parameters $\hat{\boldsymbol{\Theta}}_{\mathrm{smpl}}$ for deployment)}} \\
\midrule
\rowcolor{rowbg}
\textit{Detect}
&
$b_i \leftarrow \mathrm{Detect}(\mathbf{I})$ \hfill {\color{anno} body boxes only}
&
$b_i\,\, b_{\mathrm{L}},\, b_{\mathrm{R}} \leftarrow \mathrm{Detect}(\mathbf{I})$ \hfill {\color{anno} body + hand boxes}
\\[1pt]
\textit{Encode}
&
$\mathbf{F}_{\mathrm{body}} \leftarrow \mathrm{Enc}(\mathrm{Crop}(\mathbf{I},\, b_i))$ \hfill {\color{anno} backbone \#1}
&
$[\mathbf{F}_{\mathrm{body}},\, \mathbf{F}_{\mathrm{L}},\, \mathbf{F}_{\mathrm{R}}]$
\newline $\;\leftarrow \mathrm{Enc}(\mathrm{GPUCrop}(\mathbf{I},\, b_i,\, b_{\mathrm{L}},\, b_{\mathrm{R}}))$ \hfill {\color{anno} \textbf{backbone $\boldsymbol{\times}$1, batched}}
\\[1pt]
\rowcolor{rowbg}
\textit{Body}
&
$\hat{\theta} \leftarrow \mathrm{BodyDec}(\mathbf{F}_{\mathrm{body}})$
&
$\hat{\theta} \leftarrow \mathrm{BodyDec}(\mathbf{F}_{\mathrm{body}})$
\\[1pt]
\textit{Hands}
&
$b_{\mathrm{L}},\, b_{\mathrm{R}} \leftarrow \mathrm{ProjectWrist}(\hat{\theta})$ \hfill {\color{anno} \textbf{wait for body dec.}}
\newline $\mathbf{F}_{\mathrm{L}} \leftarrow \mathrm{Enc}(\mathrm{CPUCrop}(\mathbf{I},\, b_{\mathrm{L}}))$ \hfill {\color{anno} backbone \#2}
\newline $\mathbf{F}_{\mathrm{R}} \leftarrow \mathrm{Enc}(\mathrm{CPUCrop}(\mathbf{I},\, b_{\mathrm{R}}))$ \hfill {\color{anno} backbone \#3}
\newline $\hat{\theta}_{\mathrm{L}} \leftarrow \mathrm{HandDec}(\mathbf{F}_{\mathrm{L}})$
\newline $\hat{\theta}_{\mathrm{R}} \leftarrow \mathrm{HandDec}(\mathbf{F}_{\mathrm{R}})$ \hfill {\color{anno} sequential, 2 passes}
&
$\hat{\theta}_{\mathrm{L}},\, \hat{\theta}_{\mathrm{R}} \leftarrow \mathrm{HandDec}([\mathbf{F}_{\mathrm{L}},\, \mathbf{F}_{\mathrm{R}}])$ \hfill {\color{anno} \textbf{batched, 1 pass}}
\newline {\color{anno} (features from Encode stage)}
\\[1pt]
\rowcolor{rowbg}
\textit{Merge}
&
$\hat{\theta} \leftarrow \mathrm{Merge}(\hat{\theta},\, \hat{\theta}_{\mathrm{L}},\, \hat{\theta}_{\mathrm{R}})$
&
$\hat{\theta} \leftarrow \mathrm{Merge}(\hat{\theta},\, \hat{\theta}_{\mathrm{L}},\, \hat{\theta}_{\mathrm{R}})$
\\[1pt]
\textit{Refine}
&
$\hat{\theta} \leftarrow \mathrm{BodyDec}(\mathbf{F}_{\mathrm{body}},\; \hat{\mathbf{J}}_{2\mathrm{d}})$ \hfill {\color{anno} 2nd decoder pass}
&
{\color{anno} (skipped)}
\\[1pt]
\rowcolor{rowbg}
\textit{Convert}
&
$\hat{\boldsymbol{\Theta}}_{\mathrm{smpl}} \leftarrow \mathrm{IterFit}(\hat{\theta})$ \hfill {\color{anno} hundreds of iterations}
&
$\hat{\boldsymbol{\Theta}}_{\mathrm{smpl}} \leftarrow f_\omega(\hat{\theta})$ \hfill {\color{anno} single forward pass}
\\[3pt]
&
\multicolumn{2}{l}{\color{anno}\tiny optional: converts MHR output to SMPL for on-device deployment}
\\
\bottomrule
\end{tabular}
\vspace{-2mm}
\caption{\textbf{\textit{Inference pipeline comparison.}} Original 3DB pipeline with serial execution (left) and our accelerated variant with batched execution (right).}
\vspace{-1mm}
\label{fig:pipeline_comparison}
\end{figure}

\paragraph{Merged body--hand batching.}
Since the spatial prior provides body and hand bounding boxes simultaneously, all three crops can be prepared in parallel without waiting for the body decoder output. We concatenate them into a single batch and execute one backbone forward pass:
\begin{equation}
    [\,\mathbf{F}_{\mathrm{body}},\, \mathbf{F}_{\mathrm{L}},\, \mathbf{F}_{\mathrm{R}}\,] = \mathrm{Enc}\bigl([\,\mathbf{I}_{\mathrm{body}},\, \mathbf{I}_{\mathrm{L}},\, \mathbf{I}_{\mathrm{R}}\,]\bigr).
    \label{eq:merged_batch}
\end{equation}
The resulting feature maps are split and fed to the body and hand decoders, respectively, reducing backbone evaluations from three to one per person.

\paragraph{Operator-level optimizations.}
We apply a series of micro-optimizations throughout the pipeline, including replacing generic library calls with graph-traceable specialized operators, vectorizing per-joint loops, and caching frequently accessed parameters to eliminate redundant CPU--GPU synchronizations.

\subsubsection{Neural Kinematic Projection}
\label{sec:mhr2smpl}
The iterative MHR-to-SMPL fitting in Eq.~\eqref{eq:mhr2smpl} is the slowest single stage, requiring hundreds of optimization iterations per person. We obviate this cross-topology bottleneck with a lightweight feedforward network $f_\omega$ that directly regresses the actuatable SMPL kinematic manifold from the expressive MHR surface in a single pass.

\paragraph{Topology bridging.}
Since MHR and SMPL use different mesh topologies ($N_v^{\mathrm{mhr}} = 18{,}439$ vs.\ $N_v^{\mathrm{smpl}} = 6{,}890$ vertices), we first project the predicted MHR vertices onto the SMPL surface using precomputed barycentric coordinates:
\begin{equation}
    \tilde{\mathbf{V}} = \mathcal{B}(\mathbf{V}_{\mathrm{mhr}}) \;\in\; \mathbb{R}^{N_v^{\mathrm{smpl}} \times 3},
    \label{eq:bary}
\end{equation}
where $\mathcal{B}$ maps each SMPL vertex to its corresponding MHR triangle via stored face indices and barycentric weights. This operation is a single batched matrix multiply with no iterative optimization.

\paragraph{Network architecture and input representation.}
We subsample $V_{\mathrm{sub}}$ vertices from $\tilde{\mathbf{V}}$, subtract the centroid, and flatten the result into a vector $\mathbf{x} \in \mathbb{R}^{3V_{\mathrm{sub}}}$. The network $f_\omega$ is a three-layer MLP ($3V_{\mathrm{sub}} \!\to\! 512 \!\to\! 256 \!\to\! d_\Theta$) with ReLU activations. We set $V_{\mathrm{sub}} = 1{,}500$, yielding an input dimension of $4{,}500$ and approximately 2.5\,M parameters. The output $d_\Theta = 76$ decomposes into global orientation (3), body pose (63, covering 21 body joints; the two hand joints are handled by the hand decoder and zeroed here), and shape coefficients (10):
\begin{equation}
    \hat{\boldsymbol{\Theta}}_{\mathrm{smpl}} = f_\omega(\mathbf{x}).
    \label{eq:mlp_convert}
\end{equation}

\paragraph{Training.}
We generate training pairs $\{(\mathbf{x}_i,\, \boldsymbol{\Theta}_i^*)\}$ by running the original iterative fitting (Eq.~\eqref{eq:mhr2smpl}) on a large set of 3DB predictions, where $\boldsymbol{\Theta}_i^*$ denotes the converged SMPL parameters. We apply SMPL forward kinematics to $\hat{\boldsymbol{\Theta}}_{\mathrm{smpl}}$ to obtain the predicted mesh $\hat{\mathbf{V}}_{\mathrm{smpl}}$, and train the MLP with:
\begin{equation}
    \mathcal{L}_{\mathrm{convert}} = \lambda_v \|\hat{\mathbf{V}}_{\mathrm{smpl}} - \tilde{\mathbf{V}}\|_1 + \lambda_{\mathrm{reg}} \|\hat{\boldsymbol{\Theta}}_{\mathrm{smpl}} - \boldsymbol{\Theta}^*\|_2^2.
\end{equation}

\paragraph{Kinematic prior refinement.}
The per-frame regression output may contain anatomically implausible poses (\eg, kinematic artifacts). We train a lightweight denoising MLP on clean AMASS~\cite{amass} motion-capture sequences by adding synthetic noise to ground-truth poses and supervising the network to recover the original. At inference, the MLP takes the predicted SMPL body pose as input and projects it onto the learned manifold of natural human poses. This single-frame refinement adds negligible latency (${\sim}0.1$\,ms) and improves the plausibility of the converted output without affecting benchmark metrics.

\section{Experiments}
\label{sec:Experiments}

\subsection{Experimental Setup}

\noindent\textbf{Benchmarks and Metrics.}
We adopt the standard HMR evaluation suite comprising three 3D error
metrics---PA-MPJPE(PA)~\cite{pa_mpjpe_pck}, MPJPE~\cite{mpjpe}, and
PVE~\cite{pve}, all reported in millimeters---together with
PCK@0.05~\cite{pa_mpjpe_pck} for 2D alignment.
These are evaluated on six benchmarks that span a wide range of capture
conditions: 3DPW~\cite{3dpw}, EMDB~\cite{emdb}, RICH~\cite{rich},
Harmony4D~\cite{harmony4d}, COCO~\cite{coco}, and LSPET~\cite{lspet}.
Since 3DB natively predicts MHR parameters, we convert the 
predicted mesh to the SMPL~\cite{smpl} topology using the procedure 
described in~\cite{MHR:2025} before computing all 3D metrics on 
SMPL-annotated benchmarks.
For hand-specific evaluation, we additionally report PA-MPVPE and
F-scores (F@5, F@15) on FreiHAND~\cite{2019freihand}.

\noindent\textbf{Evaluation Protocols.}
We report results under two complementary settings.
In the \textit{Oracle} protocol, ground-truth bounding boxes isolate encoder-decoder accuracy from detection errors, following~\cite{sam3db} for controlled comparison.
In the \textit{Automatic} protocol, bounding boxes from a detector reflect end-to-end performance. 
Specifically, we evaluate the baseline 3DB with its original detector~\cite{vitdet,dinov3} and our method with our proposed decoupled pose priors~\cite{yolo11}.
We report per-crop FPS under \textit{Oracle} and per-frame FPS under \textit{Automatic} to detail speed-accuracy trade-offs.
For Harmony4D, we follow the leave-one-out split of~\cite{sam3db}.

\noindent\textbf{Implementation Details.}
All experiments are conducted on a single NVIDIA RTX 6000 Ada
Generation GPU with a 16-core AMD Ryzen Threadripper PRO 3955WX
CPU (3.9\,GHz, 32 threads), using PyTorch 2.5.1 and CUDA 11.5.
All throughput numbers are reported with a batch size of one
(single image at a time) to reflect the latency-critical setting.
The acceleration techniques described in Sec.~\ref{sec:accel} are
applied throughout; specific design choices are validated in the
ablation studies below.

\begin{figure}[!t]
    \centering
    \includegraphics[width=1\linewidth]{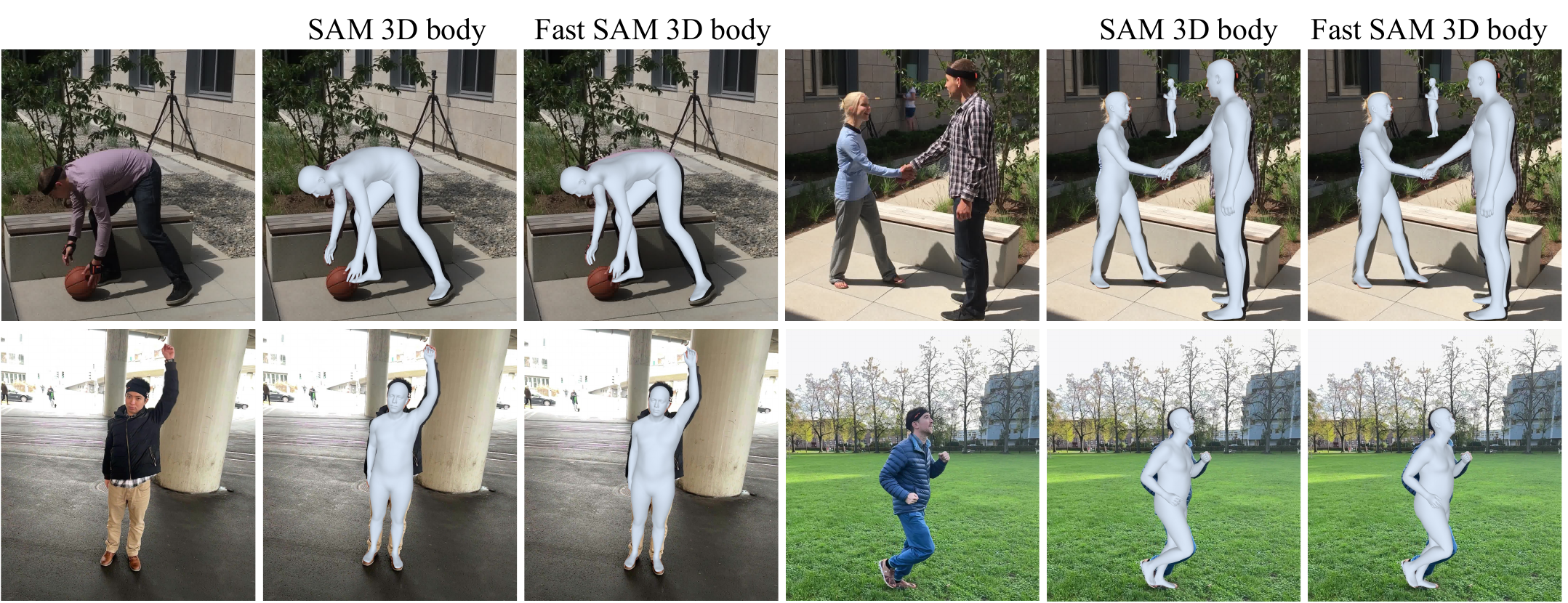}
    \vspace{-5mm}
    \caption{\textbf{\textit{Qualitative comparison.}} The original SAM 3D Body (left) and our Fast variant (right) yield visually comparable mesh reconstructions across diverse poses and multi-person scenes on 3DPW \cite{3dpw} and EMDB \cite{emdb}.}
    \vspace{-5mm}
    \label{fig:placeholder}
\end{figure}

\begin{table*}[!t]
\centering
\tablestyle{1.5pt}{1.1}
\small

\resizebox{\textwidth}{!}{%
\begin{tabular}{l|ccc|ccc|ccc|cc|c|c}
\multirow{2}{*}{Models} & \multicolumn{3}{c|}{3DPW (14)} & \multicolumn{3}{c|}{EMDB (24)} & \multicolumn{3}{c|}{RICH (24)} & \multicolumn{2}{c|}{Harmony4D (24)} & COCO & LSPET \\
\cmidrule(lr){2-4} \cmidrule(lr){5-7} \cmidrule(lr){8-10} \cmidrule(lr){11-12} \cmidrule(lr){13-13} \cmidrule(lr){14-14}
& PA$\downarrow$ & MPJPE$\downarrow$ & PVE$\downarrow$ & PA$\downarrow$ & MPJPE$\downarrow$ & PVE$\downarrow$ & PA$\downarrow$ & MPJPE$\downarrow$ & PVE$\downarrow$ & PVE$\downarrow$ & MPJPE$\downarrow$ & PCK$\uparrow$ & PCK$\uparrow$ \\
\shline
HMR2.0b       & 54.3 & 81.3 & 93.1  & 79.2 & 118.5 & 140.6 & 48.1$^\dagger$ & 96.0$^\dagger$ & 110.9$^\dagger$ & \dash & \dash & 86.1 & 53.3 \\
CameraHMR      & 35.1 & 56.0 & 65.9  & 43.3 & 70.3  & 81.7  & 34.0 & 55.7 & 64.4  & 84.6 & 70.8 & 80.5$^\dagger$ & 49.1$^\dagger$ \\
PromptHMR      & 36.1 & 58.7 & 69.4  & 41.0 & 71.7  & 84.5  & 37.3 & 56.6 & 65.5  & 91.9 & 78.0 & 79.2$^\dagger$ & 55.6$^\dagger$ \\
SMPLerX-H      & 46.6$^\dagger$ & 76.7$^\dagger$ & 91.8$^\dagger$ & 64.5$^\dagger$ & 92.7$^\dagger$ & 112.0$^\dagger$ & 37.4$^\dagger$ & 62.5$^\dagger$ & 69.5$^\dagger$ & \dash & \dash & \dash & \dash \\
NLF-L+fit$^*$  & \underline{33.6} & \underline{54.9} & \underline{63.7}  & 40.9 & 68.4  & 80.6  & \textbf{28.7}$^\dagger$ & \textbf{51.0}$^\dagger$ & \textbf{58.2}$^\dagger$ & 97.3 & 84.9 & 74.9$^\dagger$ & 54.9$^\dagger$ \\
WHAM           & 35.9 & 57.8 & 68.7  & 50.4 & 79.7  & 94.4  & \dash & \dash & \dash & \dash & \dash & \dash & \dash \\
TRAM           & 35.6 & 59.3 & 69.6  & 45.7 & 74.4  & 86.6  & \dash & \dash & \dash & \dash & \dash & \dash & \dash \\
GENMO          & 34.6 & \textbf{53.9} & 65.8  & 42.5 & 73.0  & 84.8  & 39.1 & 66.8 & 75.4  & \dash & \dash & \dash & \dash \\
\hline
\multicolumn{14}{l}{\it \textcolor{secdark}{Oracle}: \textcolor{gray}{evaluated with annotated bounding boxes}} \\
\rowcolor{cvprblue!4}
3DB               & 33.8 & 54.8 & \textbf{63.6}  & \underline{38.2} & \textbf{61.7}  & \textbf{72.5}  & \underline{30.9} & \underline{53.7} & \underline{60.3}  & \textbf{41.0} & \textbf{33.9} & \textbf{86.5} & \underline{67.8} \\
\rowcolor{cvprblue!8}
Ours              & \textbf{30.4} & 59.5 & 68.9  & \textbf{37.3} & \underline{64.3}  & \underline{74.4}  & 33.8 & 58.8 & 64.2  & \underline{46.9} & \underline{41.9} & \textbf{86.5} & \textbf{70.2} \\
\hline
\multicolumn{14}{l}{\it \textcolor{secdark}{Automatic}: \textcolor{gray}{evaluated with detected bounding boxes}} \\
\rowcolor{cvprblue!4}
3DB               & \textbf{26.8} & 57.6 & 68.5  & 36.9 & 67.9  & \textbf{71.5}  & 33.9 & \textbf{55.4} & \textbf{60.6}  & 44.1 & 37.3 & 85.1 & 66.5 \\
\rowcolor{cvprblue!8}
Ours              & 29.7 & \textbf{58.9} & \textbf{68.0}  & \textbf{36.2} & \textbf{65.3}  & 75.6  & \textbf{33.8} & 58.7 & 64.1  & \textbf{44.0} & \textbf{37.2} & \textbf{85.2} & \textbf{67.1} \\
\hline
\multicolumn{14}{l}{\it \textcolor{gray}{Throughput}: {\color{gray_base}3DB} {\color{gray_arrow}$\rightarrow$} {\color{cvprblue}\textbf{Fast}} {\color{gray_arrow}(}{\color{green_spd}\textbf{speedup$\times$}}{\color{gray_arrow})}} \\
\rowcolor{cvprblue!4}
\textcolor{secdark}{\textit{Oracle}} {\scriptsize\color{gray}per crop}
  & \multicolumn{3}{c|}{\fps{2.0}{9.0}{4.5}}
  & \multicolumn{3}{c|}{\fps{1.9}{8.4}{4.4}}
  & \multicolumn{3}{c|}{\fps{2.0}{8.3}{4.2}}
  & \multicolumn{2}{c|}{\fps{1.4}{8.7}{6.2}}
  & \fps{1.9}{9.7}{5.1}
  & \fps{1.9}{9.7}{5.1} \\
\rowcolor{cvprblue!12}
\textcolor{secdark}{\textit{Automatic}} {\scriptsize\color{gray}per frame}
  & \multicolumn{3}{c|}{\fps{0.8}{6.6}{8.3}}
  & \multicolumn{3}{c|}{\fps{1.0}{8.3}{8.3}}
  & \multicolumn{3}{c|}{\fps{0.8}{8.2}{10.3}}
  & \multicolumn{2}{c|}{\fps{0.6}{6.5}{10.9}}
  & \fps{0.8}{6.3}{7.9}
  & \fps{0.8}{7.6}{9.5} \\
\shline
\end{tabular}
}
\vspace{1.5mm}
\caption{\textbf{\textit{Comparison on common benchmarks.}}
Oracle uses ground-truth boxes; Automatic uses detected boxes.
\textbf{Bold}: best, \underline{underline}: second-best.
$\dagger$: public checkpoint. $*$: trained on RICH.
Harmony4D: leave-one-out split~\cite{sam3db}.
PA = PA-MPJPE.
}
\vspace{-7mm}
\label{tab:main}
\end{table*}

\subsection{Comparison on Common Datasets}

To comprehensively assess our pipeline, we report quantitative and qualitative results on both body reconstruction and hand-specific reconstruction.

\noindent\textbf{Body reconstruction} Table~\ref{tab:main} compares our Fast SAM 3D Body pipeline against the original SAM 3D Body (3DB)~\cite{sam3db} and a comprehensive set of recent HMR methods, including both image-based approaches (HMR2.0b~\cite{hmr2b}, CameraHMR~\cite{camerahmr}, PromptHMR~\cite{prompthmr}, SMPLer-X-H~\cite{smplerx}, NLF~\cite{nlf}) and video-based methods (WHAM~\cite{wham}, TRAM~\cite{tram}, GenMo~\cite{genmo}).

\textit{Oracle evaluation (GT bbox provided).} Our method preserves the strong reconstruction quality of the original 3DB while achieving $4$--$6\times$ higher throughput per crop. 
For instance, on 3DPW, our Fast variant accelerates the per-crop inference by over $4.5\times$ (reaching 9.0\,FPS), while incurring a marginal MPJPE increase of less than 5\,mm compared to the original 3DB.
We note that the PA-MPJPE metric, which factors out global alignment, often shows smaller degradation than MPJPE. This indicates that the simplified model maintains accurate local pose estimation, trading off only slightly in global positioning. Qualitative visualizations (Fig.~\ref{fig:placeholder}) further corroborate this preserved spatial accuracy.

\textit{Automatic evaluation (use pred bbox).}
In the Automatic setting, our method frequently matches or surpasses the 3DB baseline (\emph{e.g.}, on EMDB and Harmony4D). This confirms that extracting hand crops from coarse YOLO11-Pose~\cite{yolo11} wrist keypoints provides a sufficient spatial prior for the robust downstream decoder, requiring no sub-pixel precision. Crucially, breaking the baseline's serial dependency unlocks batched TensorRT execution, driving an $8$ to $11\times$ end-to-end throughput boost while fully preserving geometric fidelity.

\textit{2D alignment.}
On COCO and LSPET (PCK@0.05), our method achieves highly competitive results under Oracle evaluation, matching or exceeding the original 3DB and all other baselines. 
The consistent improvement on LSPET may be attributed to the simplified decoder encouraging more robust 2D projections on out-of-domain data.

\begin{figure}[!h]
\centering
\begin{minipage}[c]{0.48\linewidth}
\centering
\scriptsize
\tablestyle{1pt}{1.1}
\begin{tabular}{l|cc}
Metric & 3DB & Ours \\
\shline
PA-MPJPE$\downarrow$ & 5.50 & 5.67 \\
PA-MPVPE$\downarrow$ & 6.20 & 6.46 \\
F@5$\uparrow$        & 0.737 & 0.714 \\
F@15$\uparrow$       & 0.988 & 0.987 \\
\hline
\multicolumn{3}{l}{\it \textcolor{gray}{Throughput}: {\color{gray_base}3DB} {\color{gray_arrow}$\rightarrow$} {\color{cvprblue}\textbf{Ours}} {\color{gray_arrow}(}{\color{green_spd}\textbf{speedup$\times$}}{\color{gray_arrow})}} \\
\rowcolor{cvprblue!8}
\textit{\scriptsize FPS}
  & \multicolumn{2}{c}{\fps{9.8}{45.5}{4.6}} \\
\shline
\end{tabular}
\end{minipage}%
\hspace{4pt}%
\begin{minipage}[c]{0.4\linewidth}
\centering
\includegraphics[width=\linewidth]{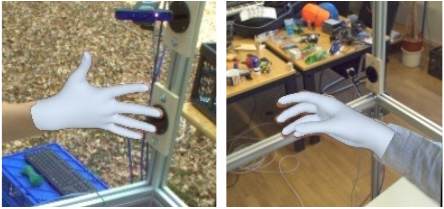}
\end{minipage}
\vspace{-3mm}
\caption{\textbf{\textit{Hand reconstruction.}}
(Left) Quantitative comparison on FreiHAND \cite{2019freihand}: our lightweight hand decoder maintains competitive accuracy with $4.6\!\times$ faster inference.
(Right) Qualitative results.}
\vspace{-5mm}
\label{fig:hand_reconstruction}
\end{figure}

\noindent\textbf{Hand reconstruction.}
Figure~\ref{fig:hand_reconstruction} evaluates our lightweight hand
decoder on FreiHAND~\cite{2019freihand}.
The simplified decoder maintains competitive accuracy across all metrics,
with differences within 0.3\,mm on PA-MPJPE/PA-MPVPE and near-identical
F@15 scores, while achieving a $4.6\times$ throughput improvement.
Qualitatively, our method recovers high-fidelity hand meshes robust to diverse poses and background clutter.

\subsection{Real-World Deployment}
Deploying on humanoid platforms requires accurate SMPL output at
real-time throughput. We address both, enabling vision-only
teleoperation of the Unitree G1 from a single RGB stream.

\begin{table}[!h]
\centering
\tablestyle{3pt}{1.1}
\scriptsize
\begin{tabular}{l|ccc|ccc}
\multirow{2}{*}{Method} & \multicolumn{3}{c|}{3DPW} & \multicolumn{3}{c}{EMDB} \\
\cmidrule(lr){2-4} \cmidrule(lr){5-7}
& PA$\downarrow$ & MPJPE$\downarrow$ & PVE$\downarrow$ & PA$\downarrow$ & MPJPE$\downarrow$ & PVE$\downarrow$ \\
\shline

(3DB) Iterative fitting  & \textbf{30.4} & 59.5 & 68.9 & \textbf{37.3} & \textbf{64.3} & 74.4 \\

(Ours) Feedforward MLP    & 31.1 & \textbf{59.4} & \textbf{68.6} & \textbf{37.3} & 64.8 & \textbf{68.8} \\
\hline
\multicolumn{7}{l}{\it \textcolor{gray}{Throughput}: {\color{gray_base}Iterative} {\color{gray_arrow}$\rightarrow$} {\color{cvprblue}\textbf{MLP}} {\color{gray_arrow}(}{\color{green_spd}\textbf{speedup$\times$}}{\color{gray_arrow})}} \\
\rowcolor{cvprblue!8}
\textit{\scriptsize FPS}
  & \multicolumn{3}{c|}{\fps{0.24}{1992}{8300}}
  & \multicolumn{3}{c}{\fps{0.23}{2398}{10426}} \\
\shline
\end{tabular}
\vspace{1mm}
\caption{\textbf{\textit{Lightweight MHR-to-SMPL conversion.}} Iterative fitting vs.\ feedforward MLP, Oracle protocol. Near-identical accuracy at $>$$10^4\!\times$ speedup.}
\label{tab:ablation_mhr2smpl}
\vspace{-8mm}
\end{table}


\noindent\textbf{Lightweight MHR-to-SMPL conversion.}
Table~\ref{tab:ablation_mhr2smpl} compares the iterative fitting (Eq.~\ref{eq:mhr2smpl}) against our feedforward MLP (Eq.~\ref{eq:mlp_convert}). The MLP achieves near-identical joint accuracy (PA-MPJPE and MPJPE) to the iterative baseline across both benchmarks, and even noticeably improves the vertex-level alignment (PVE) on EMDB. Crucially, this feedforward approach yields a speedup of roughly four orders of magnitude (${\sim}10^4\times$), effectively eliminating the final latency bottleneck in our pipeline.

\begin{figure}[!t]
    \centering
    \includegraphics[width=1\linewidth]{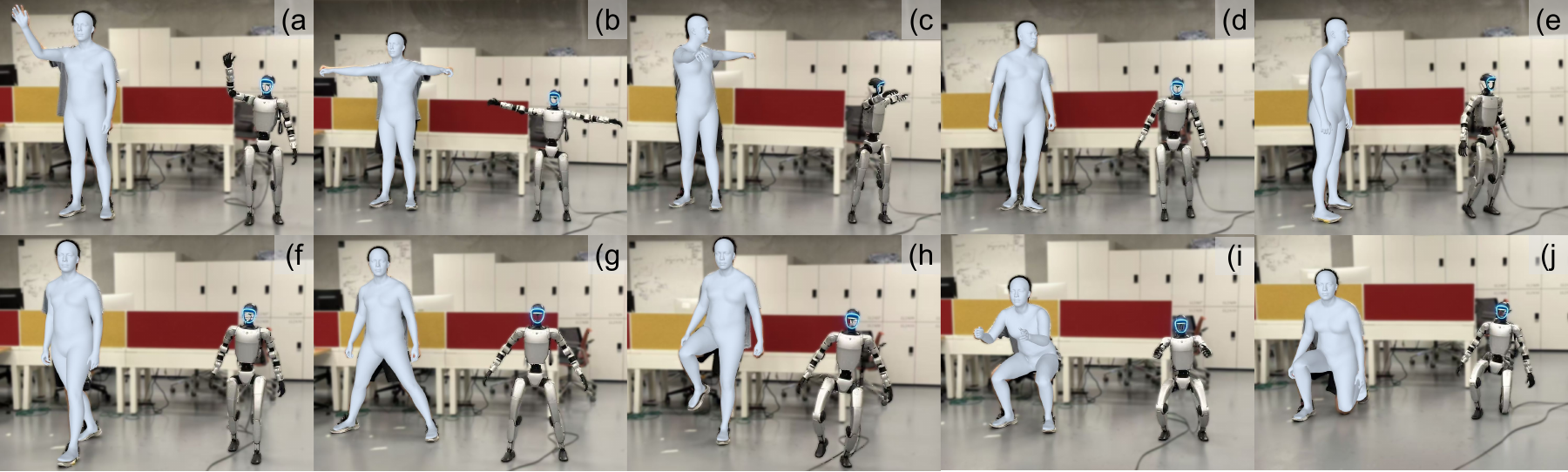}
    \vspace{-6mm}
    \caption{\textbf{\textit{Qualitative Results of Humanoid Teleoperation.}} The system tracks diverse whole-body motions including upper-body gestures~(a), body rotations~(b--e), walking~(f), wide stance~(g), single-leg standing~(h), squatting~(i), and kneeling~(j).}
    \label{fig:teleop_qual_results}
\end{figure}

\textbf{Humanoid teleoperation.} We demonstrate a real-time, vision-only teleoperation system for the Unitree G1 humanoid robot using a single RGB camera. Operating at an end-to-end latency of just 65 ms on an NVIDIA RTX 5090, our framework recovers full-body MHR parameters and instantly translates them via our neural kinematic projection (Sec.~\ref{sec:mhr2smpl}). These derived SMPL kinematics are directly fed as reference motion into SONIC~\cite{luo2025sonic} to drive the G1. Qualitative results are visualized in Fig.~\ref{fig:teleop_qual_results}. 

\begin{figure}[!t]
    \centering
    \vspace{-3mm}
    \includegraphics[width=1\linewidth]{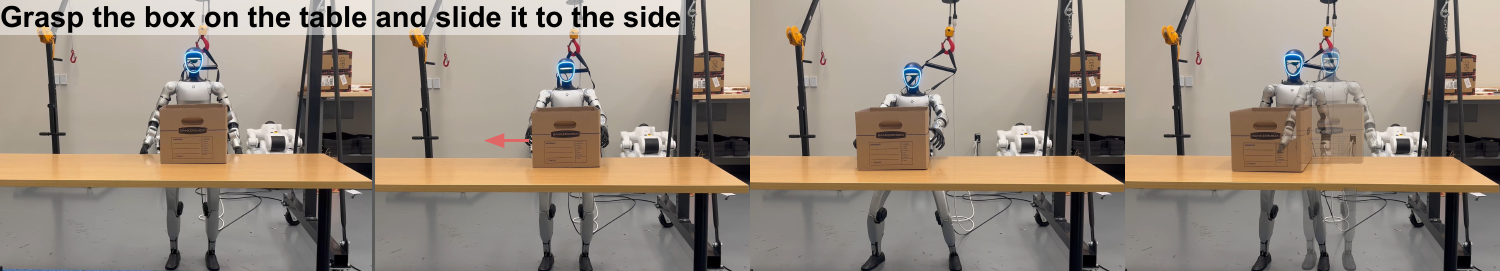}
    \caption{\textbf{\textit{Humanoid Policy Rollout.}} The robot grasps a box on the table with both hands, squats down, and steps to the right.}
    \vspace{-6mm}
    \label{fig:teleop_policy_results}
\end{figure}

We further validate our pipeline by finetuning $\Psi_{0}$~\cite{wei2026psi0}—a VLA model pre-trained on diverse human~\cite{hoque2025egodex} and humanoid~\cite{zhao2025humanoideverydaycomprehensiverobotic} priors—using 40 demonstrations collected via our system. Evaluated on a whole-body manipulation task (bimanual grasping, squatting, and lateral stepping), the policy achieves an 80\% success rate (Fig.~\ref{fig:teleop_policy_results}). This confirms our framework generates high-quality motion data suitable for deployable policy learning.

\subsection{Ablation Studies}

We conduct extensive ablations to justify each design choice, evaluating
on 3DPW with oracle GT bounding boxes unless otherwise noted.

\begin{table}[!h]
\centering
\begin{minipage}[t]{0.62\linewidth}
\centering
\tablestyle{1pt}{1.1}
\scriptsize

\resizebox{\linewidth}{!}{%
\begin{tabular}{l|cc|c|c||l|cc|c|c}
\multicolumn{5}{c||}{\it \textcolor{secdark}{Multi-layer}} & \multicolumn{5}{c}{\it \textcolor{secdark}{Single-layer}} \\
\cmidrule(lr){1-5} \cmidrule(lr){6-10}
Layers & MPJPE$\downarrow$ & PA$\downarrow$ & PVE$\downarrow$ & FPS$\uparrow$ & Layer & MPJPE$\downarrow$ & PA$\downarrow$ & PVE$\downarrow$ & FPS$\uparrow$ \\
\shline
\rowcolor{cvprblue!8}
\{0,1,2\} & \underline{58.96} & \underline{31.33} & \underline{69.28} & 7.18 & L0 & 60.17 & 33.76 & \textbf{70.26} & 7.52 \\
\{0,1,2,3,4\}        & \textbf{58.89} & \textbf{31.21} & \textbf{69.21} & 6.88 & L1 & 60.03 & 33.41 & 71.27 & 7.65 \\
\{0,2,4\}            & 59.00 & 31.43 & 69.53 & 7.18 & L2 & \textbf{59.15} & \textbf{32.62} & \underline{71.21} & 7.71 \\
\{0,4\}              & 60.23 & 33.67 & 70.22 & 7.31 & L3 & \underline{59.89} & 33.04 & 72.59 & 7.88 \\
$\varnothing$        & 60.73 & 33.38 & 72.41 & \textbf{8.19} & L4 & 60.16 & \underline{33.03} & 71.86 & \textbf{7.95} \\
\shline
\end{tabular}
}
\vspace{1mm}
\caption{\textbf{\textit{Ablation on intermediate prediction layers
for the body decoder.}} Multi-layer (left) retains a subset of the
five original layers; single-layer (right) keeps only one.
Hand decoder layers fixed at \{0,1\}. \colorbox{cvprblue!8}{Shaded rows} indicate selected configuration.}

\label{tab:ablation_layers}
\end{minipage}
\hfill
\begin{minipage}[t]{0.35\linewidth}
\centering
\tablestyle{1pt}{1.1}
\scriptsize

\resizebox{\linewidth}{!}{%
\begin{tabular}{l|cc|c|c}
Size & MPJPE$\downarrow$ & PA$\downarrow$ & PVE$\downarrow$ & FPS$\uparrow$ \\
\shline
128  & 86.78 & 62.33 & 129.10 & \textbf{11.69} \\
256  & 61.70 & 33.45 & 75.33 & 10.73 \\
384  & 59.01 & \textbf{31.00} & 70.22 & 8.96 \\
448  & \textbf{58.95} & 31.02 & 69.50 & 8.25 \\
\rowcolor{cvprblue!8}
512 & 58.96 & 31.33 & \textbf{69.28} & 7.18 \\
\shline
\end{tabular}

}
\vspace{1mm}
\caption{\textbf{\textit{Input resolution ablation.}} Smaller crops
improve throughput at the cost of reconstruction accuracy.
\colorbox{cvprblue!8}{Shaded row} indicates selected configuration.}
\label{tab:ablation_resolution}
\vspace{-5mm}
\end{minipage}
\end{table}

\noindent\textbf{Decoder layer selection.}
Table~\ref{tab:ablation_layers} evaluates the number of transformer layers in the body decoder. 
The multi-layer configuration $\{0,1,2\}$ achieves the best accuracy--speed trade-off, effectively matching the full model's reconstruction quality while delivering a noticeable boost in inference speed. 
Further reducing the number of layers leads to clear precision loss. 
While single-layer options offer higher throughput, they fail to maintain the robust performance of our selected multi-layer setting.

\begin{table}[!h]
\centering
\begin{minipage}[t]{0.48\linewidth}
\centering
\tablestyle{1pt}{1.1}
\scriptsize

\begin{tabular}{l|c|cc|c|c}
Model & Setting & MPJPE$\downarrow$ & PA$\downarrow$ & PVE$\downarrow$ & FPS$\uparrow$ \\
\shline
\multicolumn{6}{l}{\it \textcolor{secdark}{(a) Model size}} \\
\rowcolor{cvprblue!8}
small {\scriptsize(TRT)} & 35M & 59.08 & 31.33 & 69.33 & 7.34 \\
\rowcolor{cvprblue!8}
small           & 35M & 59.04 & 31.34 & 69.32 & \textbf{7.38} \\
base                       & 104M & 59.06 & 31.35 & 69.35 & 7.31 \\
large                       & 331M & \textbf{58.95} & \textbf{31.32} & \textbf{69.23} & 7.24 \\
\hline
\multicolumn{6}{l}{\it \textcolor{secdark}{(b) Resolution level}} \\
\rowcolor{cvprblue!8}
0  & ${\sim}$1200 & 59.08 & 31.34 & 69.33 & \textbf{7.45} \\
5            & ${\sim}$2400 & 58.95 & 31.33 & \textbf{69.18} & 7.42 \\
9            & ${\sim}$3600 & \textbf{58.95} & \textbf{31.32} & 69.23 & 7.38 \\
\shline
\end{tabular}
\vspace{1mm}
\caption{\textbf{\textit{FOV estimator ablation.}}
(a) Model size has negligible impact on accuracy (params).
(b) Resolution has diminishing returns (tokens).
\colorbox{cvprblue!8}{Shaded rows} indicate selected configuration.}
\vspace{-5mm}
\label{tab:ablation_fov}

\end{minipage}
\hfill
\begin{minipage}[t]{0.48\linewidth}
\centering
\tablestyle{1pt}{1.1}
\scriptsize

\begin{tabular}{l|l|cc|c|c}
Component & Setting & MPJPE$\downarrow$ & PA$\downarrow$ & PVE$\downarrow$ & FPS$\uparrow$ \\
\shline
\multicolumn{6}{l}{\it \textcolor{secdark}{(a) Pose-dependent Blend Shapes}} \\
\rowcolor{cvprblue!8}
Correctives & OFF & 58.96 & 31.33 & 69.28 & \textbf{7.18} \\
Correctives & ON             & \textbf{58.04} & \textbf{31.00} & \textbf{69.26} & 6.89 \\
\hline
\multicolumn{6}{l}{\it \textcolor{secdark}{(b) Keypoint Prompt}} \\
\rowcolor{cvprblue!8}
KP Prompt & OFF   & 58.96 & \textbf{31.33} & \textbf{69.28} & \textbf{7.18} \\
KP Prompt & ON               & \textbf{58.92} & 31.34 & 70.96 & 6.07 \\
\hline
\multicolumn{6}{l}{\it \textcolor{secdark}{(c) Corr.\ + KP combined}} \\
\rowcolor{cvprblue!8}
Corr.+KP & OFF & 58.96 & 31.33 & \textbf{69.28} & \textbf{7.18} \\
Corr.+KP & ON  & \textbf{58.00} & \textbf{31.01} & 70.94 & 5.78 \\
\shline
\end{tabular}
\vspace{1mm}
\caption{\textbf{\textit{Component ablations.}} Each section toggles
one component from the Fast baseline. (c) shows the cumulative effect
of all simplifications.
\colorbox{cvprblue!8}{Shaded rows} indicate selected configuration.}
\vspace{-5mm}
\label{tab:ablation_components}

\end{minipage}
\end{table}

\noindent\textbf{Input resolution.}
Table~\ref{tab:ablation_resolution} demonstrates that input resolution significantly impacts performance. 
While accuracy saturates at higher resolutions, the 384/448-pixel variant provides a highly efficient alternative with minimal quality loss. 
We maintain the maximum resolution as the default to remain consistent with the original 3DB baseline, but note that the mid-range option is ideal for latency-sensitive applications.

\noindent\textbf{Pose-dependent correctives and keypoint prompts.}
Table~\ref{tab:ablation_components} evaluates the removal of the 3DB baseline's heavy refinement modules. Disabling pose-dependent correctives and keypoint prompts significantly accelerates inference, as these components primarily govern high-frequency surface details rather than joint-level kinematics. These results confirm that our streamlined architecture — relying on decoupled spatial priors derived from a lightweight pose detector rather than heavy iterative prompting—delivers a highly efficient alternative with negligible impact on overall reconstruction quality.

\noindent\textbf{FOV estimator.}
Table~\ref{tab:ablation_fov} ablates the MoGe-2~\cite{moge2} field-of-view estimator across model sizes and input resolutions. 
We find that accuracy remains nearly constant across all model scales, suggesting that the FOV estimation task is well-saturated and the smallest model is sufficient. 
Similarly, increasing the input resolution yields negligible accuracy gains while slightly reducing throughput. 
Consequently, we adopt the small model at the lowest resolution to maximize speed without sacrificing performance.

\begin{table}[!h]
\centering
\tablestyle{1pt}{1.1}
\scriptsize
\begin{tabular}{l|l|c|cc|c|c}
Detector & Setting & Recall & MPJPE$\downarrow$ & PA$\downarrow$ & PVE$\downarrow$ & FPS$\uparrow$ \\
\shline
\multicolumn{7}{l}{\it \textcolor{secdark}{(a) Detector type}} \\
\rowcolor{cvprblue!8}
YOLO11m-Pose & body + hand & 99.2\% & \textbf{58.49} & \textbf{30.71} & \textbf{67.36} & \textbf{3.50} \\
YOLO11m (det only)       & body only & 99.7\% & 58.81 & 30.97 & 67.71 & 2.60 \\
ViTDet-H                 & body only & \textbf{100\%} & 58.79 & 30.98 & 67.70 & 1.05 \\
\hline
\multicolumn{7}{l}{\it \textcolor{secdark}{(b) YOLO-Pose model size}} \\
\rowcolor{cvprblue!8}
YOLO11n-Pose  & 3M  & 98.3\% & \textbf{58.30} & \textbf{30.54} & \textbf{67.06} & \textbf{4.15} \\
YOLO11s-Pose  & 10M & 99.0\% & 58.47 & 30.75 & 67.31 & 3.56 \\
\rowcolor{cvprblue!8}
YOLO11m-Pose & 20M & \textbf{99.3\%} & 58.49 & 30.71 & 67.36 & 3.50 \\
YOLO11l-Pose  & 26M & \textbf{99.3\%} & 58.57 & 30.77 & 67.41 & 3.47 \\
YOLO11x-Pose  & 59M & \textbf{99.3\%} & 58.63 & 30.85 & 67.47 & 3.38 \\
\hline
\multicolumn{7}{l}{\it \textcolor{secdark}{(c) TRT vs PyTorch}} \\
\rowcolor{cvprblue!8}
YOLO11m-Pose & PyTorch & 99.2\% & 58.49 & \textbf{30.71} & 67.36 & 3.50 \\
YOLO11m-Pose             & TensorRT FP16 & 99.2\% & \textbf{58.48} & 30.72 & \textbf{67.32} & \textbf{3.58} \\
\shline
\end{tabular}
\vspace{1mm}
\caption{\textbf{\textit{Person detector ablation (automatic mode).}}
(a) YOLO11-Pose replaces ViTDet-H with comparable accuracy at higher throughput.
(b) Larger model sizes yield diminishing returns beyond medium.
(c) TensorRT conversion provides a modest additional speedup.
\colorbox{cvprblue!8}{Shaded rows} indicate the selected configurations.}
\vspace{-4mm}
\label{tab:ablation_detector}
\end{table}

\begin{table}[!h]
\centering
\tablestyle{1pt}{1.1}
\scriptsize
\begin{tabular}{l|l|cc|c|c}
Batch Mode & Description & MPJPE$\downarrow$ & PA$\downarrow$ & PVE$\downarrow$ & FPS$\uparrow$ \\
\shline
\rowcolor{cvprblue!8}
full\_batch & Body + Hand batched & 57.81 & 30.97 & 70.85 & \textbf{5.28} \\
hand\_batch            & Hand only batched   & 57.75 & 30.97 & 70.79 & 5.20 \\
no\_batch              & No batching         & \textbf{57.67} & \textbf{30.96} & \textbf{70.73} & 4.39 \\
\shline
\end{tabular}
\vspace{1mm}
\caption{\textbf{\textit{Batch mode ablation.}} Full batching
(body + hand crops in a single forward pass) improves throughput
with negligible accuracy overhead.
\colorbox{cvprblue!8}{Shaded row} indicates the selected configuration.}
\vspace{-5mm}
\label{tab:ablation_batch}
\end{table}

\noindent\textbf{Person detector.}
Table~\ref{tab:ablation_detector} evaluates detector choices under the Automatic protocol. 
YOLO11m-Pose~\cite{yolo11} achieves high recall while natively providing both body and hand bounding boxes, eliminating the need for a separate hand detector. 
Compared to the heavyweight ViTDet-H~\cite{vitdet}, our selected detector offers a substantial speedup with negligible impact on recall and downstream accuracy. 
We also find that larger model variants offer diminishing returns, with increased parameter counts failing to improve performance. 
Finally, TensorRT conversion provides a modest additional throughput gain with no loss in accuracy.

\noindent\textbf{Batching strategy.}
Table~\ref{tab:ablation_batch} compares different batching strategies. 
Full batching, which processes body and hand crops in a single forward pass, yields a significant throughput improvement compared to processing them individually. 
The accuracy overhead introduced by full batching is negligible, confirming that shared backbone computation across different crops does not introduce significant interference. 
Therefore, we adopt full batching as our default strategy to maximize inference speed.

\section{Conclusion}
\label{sec:Conclusion}

We present Fast SAM 3D Body, a training-free framework that holistically reformulates the SAM 3D Body pipeline for real-time human mesh recovery. Our optimizations deliver up to a $10.9\times$ end-to-end speedup while strictly preserving the baseline's high-fidelity reconstruction. We further introduce a lightweight neural kinematic projection that replaces the iterative MHR-to-SMPL bottleneck, achieving over $10{,}000\times$ faster topological translation. We validate the practical impact of these latency gains by demonstrating real-time, vision-only teleoperation of the humanoid robot directly from a single RGB stream.

\section{Acknowledge}

The USC Physical Superintelligence Lab acknowledges generous supports from Toyota Research Institute, Dolby, Google DeepMind, Capital One, Nvidia, Bosch, NSF, and Qualcomm. Yue Wang is also supported by a Powell Research Award.
We thank Songlin Wei and Boqian Li for their assistance with real-world policy learning and evaluation.

\clearpage  


%
%
\bibliographystyle{splncs04}
\bibliography{main}

\clearpage
\appendix
\begin{center}
{\Large \bf Supplementary Material for Fast SAM 3D Body}
\end{center}

\section{Experiment Configuration Details}
\label{sec:appendix_configs}

\noindent All ablations evaluate on 3DPW with oracle GT boxes,
\texttt{frame\_step=10}, $n{=}2{,}000$ unless otherwise noted.
Hardware: single NVIDIA RTX\,6000 Ada, PyTorch\,2.5.1, CUDA\,11.5,
batch size\,=\,1.

\subsection{Main Comparison}
\label{sec:cfg_t1t2}

\begin{sloppypar}
Table~\ref{tab:cfg_t1t2} summarizes the inference configuration shared across all
runs for Tables~\ref{tab:main} in the main paper. The left column lists settings applied universally: body intermediate prediction layers are set to $\{0,1,2\}$, hand intermediate predictions are disabled ($\emptyset$), pose-dependent correctives and keypoint prompts are both turned off, the FOV estimator uses the small model at level~0, and both parallel decoders and hand batch merging are enabled. The right column specifies protocol-dependent settings: the \textit{Oracle} protocol uses ground-truth bounding boxes, while the \textit{Automatic} protocol uses predicted boxes 
from \texttt{yolo11n-pose}. Input resolution is $384$\,px across all datasets, with the exception of Harmony4D which uses $448$\,px.
\end{sloppypar}

\begin{table}[h!]
\centering
\tablestyle{3pt}{1.0}
\scriptsize
\begin{tabular}{@{}p{3.5cm} p{1.3cm} p{1.9cm} >{\raggedright\arraybackslash}p{3.9cm}@{}}
\toprule
\multicolumn{2}{l}{\textbf{Shared config (Ours, all datasets)}}
& \multicolumn{2}{l}{\textbf{Protocol-specific settings}} \\
\midrule
\rowcolor{rowbg}
\texttt{BODY\_INTERM\_PRED\_LAYERS} & \texttt{0,1,2}
  & \textit{Oracle}      & GT boxes;  \\
\texttt{HAND\_INTERM\_PRED\_LAYERS} & $\emptyset$
  & \textit{Automatic}   & Pred boxes; \texttt{yolo11n-pose} \\
\rowcolor{rowbg}
\texttt{MHR\_NO\_CORRECTIVES}       & \texttt{1}
  & \texttt{IMG\_SIZE}   & 384\,px (448\,px for Harmony4D) \\
\texttt{SKIP\_KEYPOINT\_PROMPT}     & \texttt{1} & & \\
\rowcolor{rowbg}
\texttt{FOV\_MODEL / LEVEL}   & \texttt{small / 0} & & \\
\texttt{PARALLEL\_DECODERS}         & \texttt{1} & & \\
\rowcolor{rowbg}
\texttt{MERGE\_HAND\_BATCH}         & \texttt{1} & & \\
\bottomrule
\end{tabular}
\caption{Inference configuration for Tables~\ref{tab:main} of the main paper.}
\label{tab:cfg_t1t2}
\end{table}

\subsection{Tables~3--8 \& Figure~5 --- Configs}
\label{sec:cfg_ablations}

Table~\ref{tab:cfg_ablations} lists the fixed configs for each ablation in the main
paper; only the one swept parameter (\textit{var.}) differs across rows within each
table.
Table~\ref{tab:ablation_layers} sweeps body layers over
  $\{0,1,2\}$, $\{0,1,2,3,4\}$, $\{0,2,4\}$, $\{0,4\}$, $\emptyset$ (multi-layer)
  and L0--L4 individually (single-layer).
Table~\ref{tab:ablation_resolution} sweeps \texttt{IMG\_SIZE} $\in \{128, 256, 384, 448, 512\}$\,px.
Table~\ref{tab:ablation_fov} sweeps (a) \texttt{FOV\_MODEL} $\in \{\texttt{s},\texttt{b},\texttt{l}\}$
  with \texttt{FOV\_LEVEL=9} (TensorRT on small only) and
  (b) \texttt{FOV\_LEVEL} $\in \{0,5,9\}$ with model fixed to \texttt{s}.
Table~\ref{tab:ablation_components} toggles correctives (\texttt{MHR\_NO\_CORRECTIVES}) and
  keypoint prompt (\texttt{SKIP\_KEYPOINT\_PROMPT}) individually in (a),(b) and jointly in (c).
Table~\ref{tab:ablation_detector} evaluates (a) detector type: YOLO11m-Pose vs.\ YOLO11m vs.\ ViTDet-H;
  (b) YOLO-Pose size: n/s/m/l/x; (c) PyTorch vs.\ TensorRT FP16.
Table~\ref{tab:ablation_batch} compares three batching strategies using the 
full (non-pruned) baseline to isolate batching gains independently: 
\texttt{full\_batch}, which enables both parallel encoding and hand batch merging; 
\texttt{hand\_batch}, which enables hand batch merging only; 
and \texttt{no\_batch}, which disables both.
Table~\ref{tab:cfg_fig5} covers the FreiHAND hand reconstruction in 
Figure~\ref{fig:hand_reconstruction}.

\begin{table}[h!]
\centering
\tablestyle{2.5pt}{1.0}
\scriptsize
\begin{tabular}{@{}lcccccc@{}}
\toprule
\rowcolor{headerbg}
\textbf{Config key} & \textbf{T\ref{tab:ablation_layers}} & \textbf{T\ref{tab:ablation_resolution}} & \textbf{T\ref{tab:ablation_fov}} & \textbf{T\ref{tab:ablation_components}} & \textbf{T\ref{tab:ablation_detector}} & \textbf{T\ref{tab:ablation_batch}} \\
\midrule
\rowcolor{rowbg}
\texttt{BODY\_INTERM\_PRED\_LAYERS}
  & \textit{var.} & \texttt{0,1,2} & \texttt{0,1,2} & \texttt{0,1,2} & \texttt{0,1,2} & \texttt{0,..,4} \\
\texttt{HAND\_INTERM\_PRED\_LAYERS}
  & \texttt{0,1}  & \texttt{0,1}   & \texttt{0,1}   & \texttt{0,1}   & \texttt{0,1}   & \texttt{0,..,4} \\
\rowcolor{rowbg}
\texttt{IMG\_SIZE}
  & \texttt{512}  & \textit{var.}  & \texttt{512}   & \texttt{512}   & \texttt{512}   & \texttt{512}   \\
\texttt{MHR\_NO\_CORRECTIVES}
  & \texttt{1}    & \texttt{1}     & \texttt{1}     & \textit{var.}  & \texttt{1}     & \texttt{0}     \\
\rowcolor{rowbg}
\texttt{SKIP\_KEYPOINT\_PROMPT}
  & \texttt{1}    & \texttt{1}     & \texttt{1}     & \textit{var.}  & \texttt{1}     & \texttt{0}     \\
\texttt{FOV\_MODEL}
  & \texttt{s}    & \texttt{s}     & \textit{var.}  & \texttt{s}     & \texttt{s}     & \texttt{s}     \\
\rowcolor{rowbg}
\texttt{FOV\_TRT}
  & \texttt{1}    & \texttt{1}     & \textit{var.}  & \texttt{1}     & \texttt{1}     & \texttt{1}     \\
\texttt{FOV\_LEVEL}
  & \texttt{0}    & \texttt{0}     & \textit{var.}  & \texttt{0}     & \texttt{0}     & \texttt{0}     \\
\rowcolor{rowbg}
\texttt{DETECTOR}
  & \textemdash   & \textemdash    & \textemdash    & \textemdash    & \textit{var.}  & \textemdash    \\
\texttt{PARALLEL\_DECODERS}
  & \texttt{1}    & \texttt{1}     & \texttt{1}     & \texttt{1}     & \texttt{1}     & \textit{var.}  \\
\rowcolor{rowbg}
\texttt{MERGE\_HAND\_BATCH}
  & \texttt{1}    & \texttt{1}     & \texttt{1}     & \texttt{1}     & \texttt{1}     & \textit{var.}  \\
\midrule
Eval mode & oracle & oracle & oracle & oracle & \textbf{auto} & oracle \\
\bottomrule
\end{tabular}
\caption{Fixed configs per ablation table; \textit{var.}\ marks the swept parameter.}
\label{tab:cfg_ablations}
\end{table}

\begin{table}[h!]
\centering
\tablestyle{4pt}{1.0}
\scriptsize
\begin{tabular}{@{}lll@{}}
\toprule
\rowcolor{headerbg}
\textbf{Config key} & \textbf{3DB} & \textbf{Ours} \\
\midrule
\rowcolor{rowbg}
\texttt{BODY\_INTERM\_PRED\_LAYERS} & \texttt{0,1,2,3,4}        & \texttt{0,1,2} \\
\texttt{HAND\_INTERM\_PRED\_LAYERS} & \texttt{0,1,2,3,4}   & \texttt{0,1} \\
\rowcolor{rowbg}
\texttt{IMG\_SIZE}                  & \texttt{512\,px}         & \texttt{384\,px} \\
\texttt{MHR\_NO\_CORRECTIVES}       & \texttt{0}           & \texttt{1} \\
\rowcolor{rowbg}
\texttt{SKIP\_KEYPOINT\_PROMPT}     & \texttt{0}          & \texttt{1} \\
\bottomrule
\end{tabular}
\caption{Config for Figure~\ref{fig:hand_reconstruction} (FreiHAND). GT hand crops, oracle protocol.}
\label{tab:cfg_fig5}
\end{table}

\subsection{Throughput Breakdown}
\label{sec:speed}

\begin{figure}[!t]
    \centering
    \includegraphics[width=0.6\linewidth]{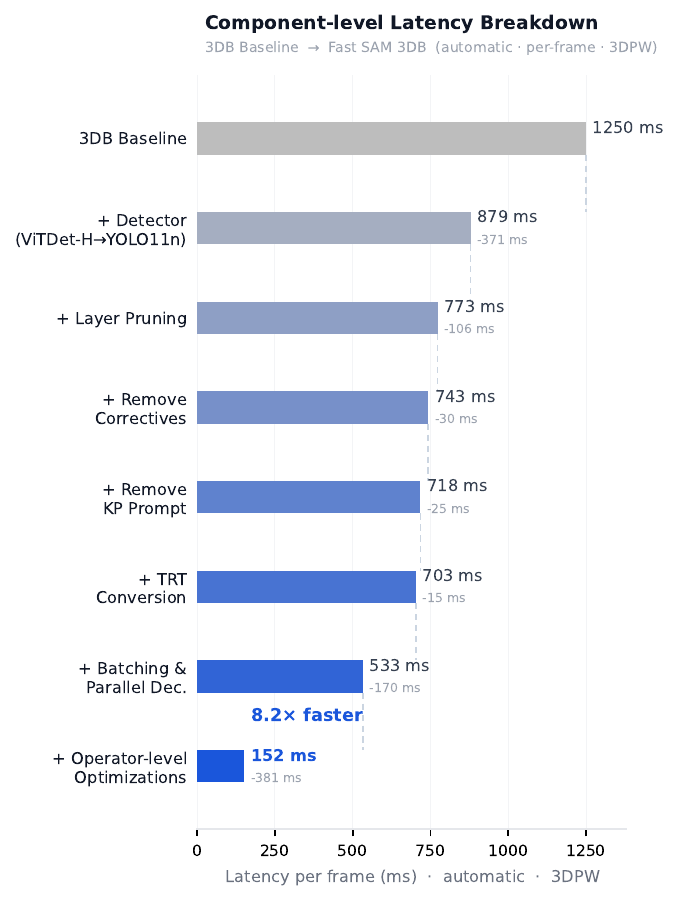}
    \caption{%
      Component-level latency breakdown from 3DB Baseline to Fast SAM\,3DB
      (automatic protocol, per-frame, 3DPW).
      Each bar shows the cumulative per-frame latency after applying the
      corresponding optimization, with the reduction annotated on the right.
      Together, the seven components yield an $8.2\times$ end-to-end speedup.%
    }
    \label{fig:throughput}
\end{figure}

\noindent Figure~\ref{fig:throughput} traces the per-component latency reduction
from the 3DB Baseline to Fast SAM\,3DB under the automatic, per-frame evaluation
protocol on 3DPW. Each step accumulates as follows.

As illustrated in Figure~\ref{fig:throughput}, while transitioning to \texttt{yolo11n-pose} saves $371$\ ms, demonstrating our system's robustness to using inferred, coarse bounding boxes rather than relying on a heavy, dedicated bounding box head—the most substantial gains stem from our core structural improvements. Specifically, our custom operator-level optimizations yield the largest single-step reduction ($381$\,ms), which, combined with parallel decoding, batching, and layer pruning, drives the overall $8.2\times$ speedup to achieve $152$\,ms, ms per frame.

\section{Qualitative Results}
\label{sec:appendix_qualitative}

\begin{figure}[!h]
    \centering
    \includegraphics[width=\linewidth]{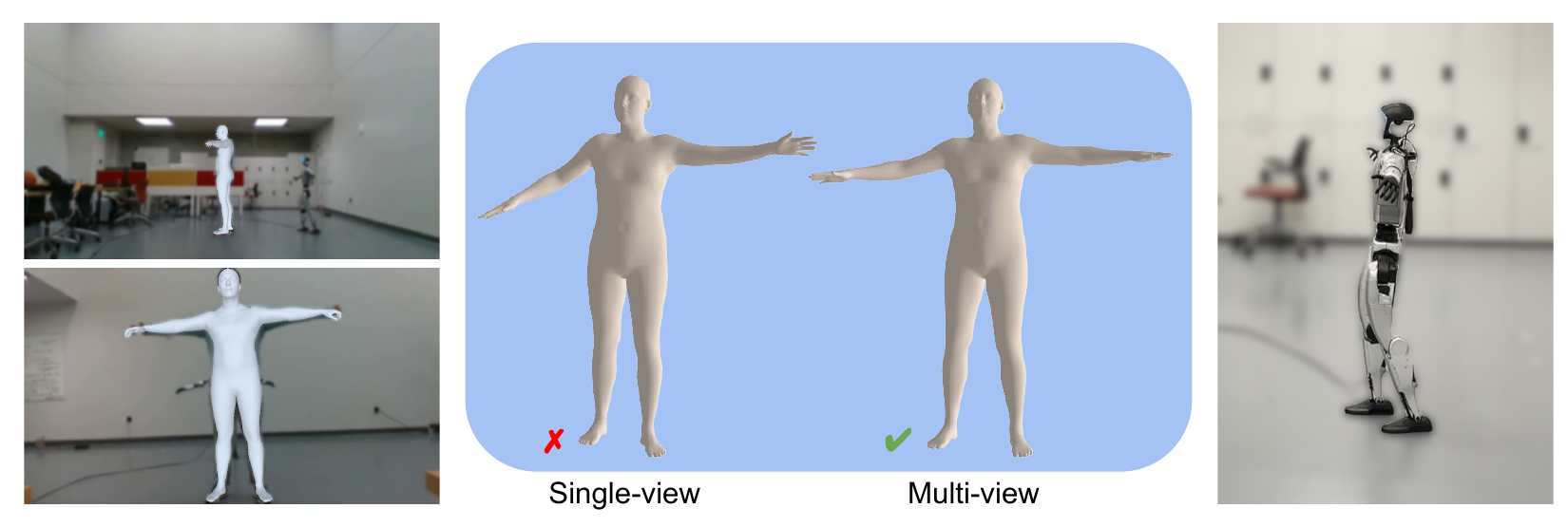}
    \caption{\textit{\textbf{Multi-view qualitative results.}}}
    \label{fig:supp_multiview}
\end{figure}

We present qualitative results of our pipeline on the 3DPW and EMDB datasets.
Each figure shows the full inference pipeline from left to right:
the original input image,
2D pose detection via YOLO-Pose,
predicted hand bounding boxes (red/blue for left/right),
the estimated body skeleton with joint connections,
the recovered mesh overlaid on the original image,
and a side-view rendering of the reconstructed 3D body mesh.
These visualizations demonstrate that our method produces plausible full-body reconstructions across diverse outdoor scenes, indoor environments, and varying numbers of people.

\begin{figure}[h!]
    \centering
    \includegraphics[width=\linewidth]{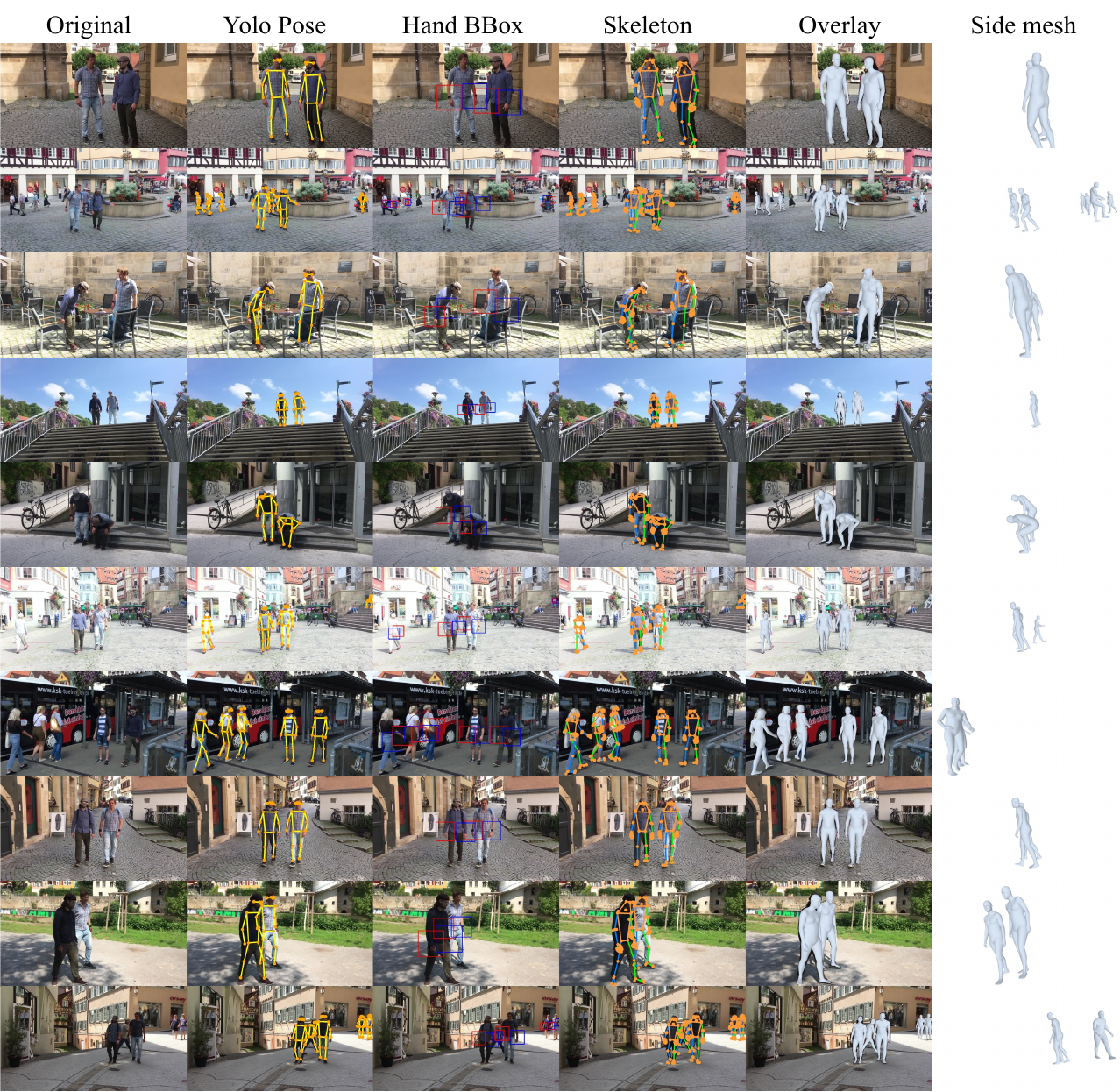}
    \vspace{-5mm}
    \caption{\textit{\textbf{Qualitative results on 3DPW}}~\cite{3dpw}. Each row shows a different scene. From left to right: original image, YOLO-Pose 2D keypoints, hand bounding box predictions, skeleton overlay, mesh overlay on the input image, and side-view 3D mesh rendering. Our method handles multi-person scenes with varying poses and occlusions.}
    \label{fig:supp_vis_3dpw}
\end{figure}

\begin{figure}[h!]
    \centering
    \includegraphics[width=\linewidth]{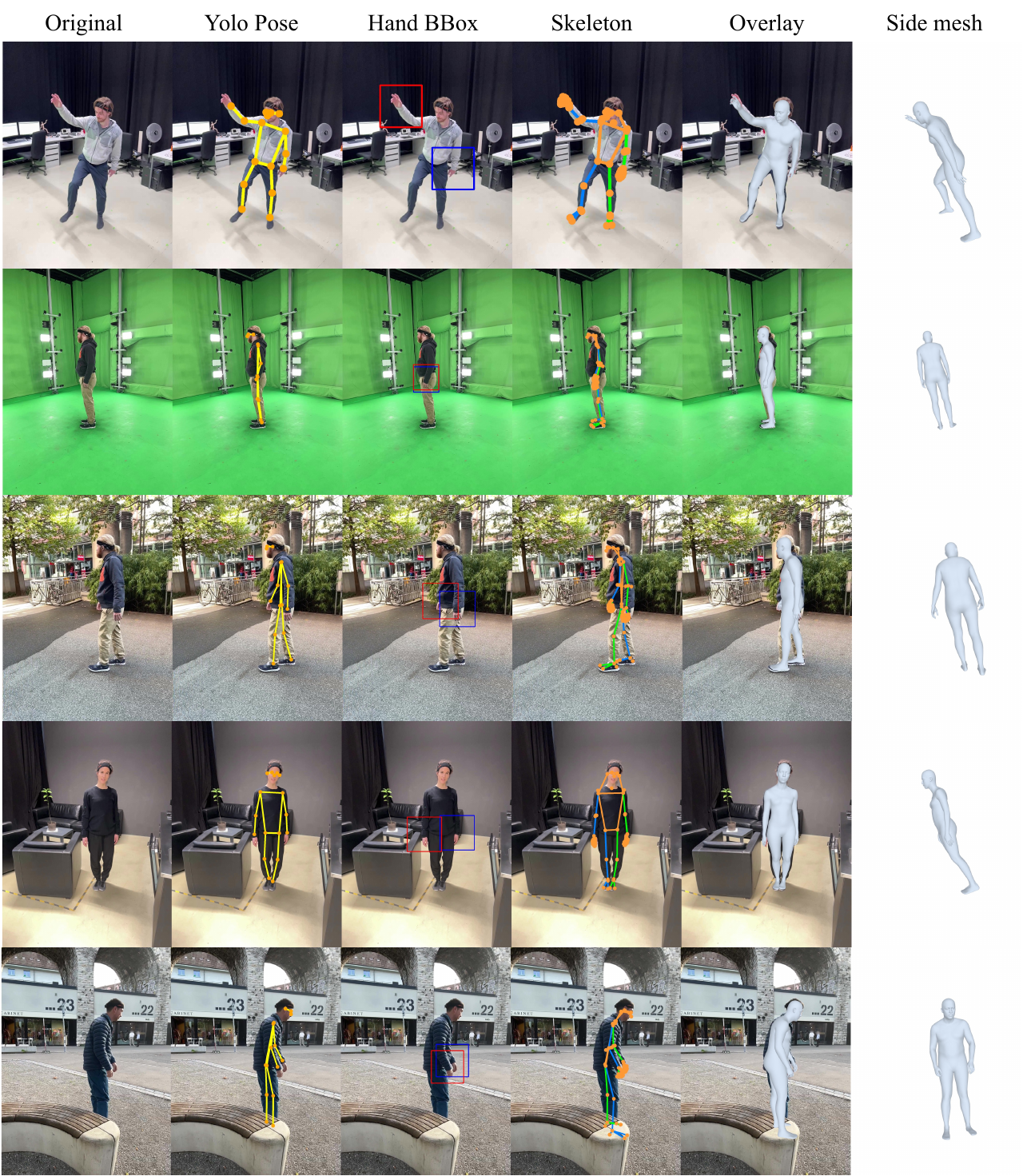}
    \vspace{-5mm}
    \caption{\textit{\textbf{Qualitative results on EMDB}}~\cite{emdb}. Each row shows a different sequence. From left to right: original image, YOLO-Pose 2D keypoints, hand bounding box predictions, skeleton overlay, mesh overlay on the input image, and side-view 3D mesh rendering. Our method generalizes well to single-person scenarios with dynamic motions in both indoor and outdoor settings.}
    \label{fig:supp_vis_emdb}
\end{figure}

\subsection{Multi-View Humanoid Teleoperation}
\label{sec:appendix_multiview}

\noindent
Self-occlusion from a single viewpoint destabilizes MHR estimates
during large body rotations, causing jitter on the robot.
To mitigate this, we extend our pipeline to a multi-view setup.
In our real-world deployment we use two cameras, while the formulation
extends to any number.

Multi-view 3D human pose estimation pipeline: images from multiple RGB cameras are batch-processed by SAM-3D-Body to extract MHR meshes, then each view's features are batched encoded into 256-dimensional vectors via a shared-weight Encoder and fused through confidence-weighted summation. The fused features are decoded into rotation (6D representation), pose, and shape parameters, which are then passed through SMPL forward kinematics to produce 3D coordinates for 24 joints. 
See Fig.~\ref{fig:supp_multiview} for qualitative results.

\end{document}